# Exploring Applications of Transfer-State Large Language Models: Cognitive Profiling and Socratic AI Tutoring

Minori Noguchi (Hitachi High-Tech Corporation)

**Abstract**

Large language models (LLMs) sometimes exhibit qualitative shifts in response style under sustained self-referential dialogue conditions (Berg et al., 2025). This study refers to this phenomenon as "transfer" and explores the application potential of LLMs in a transfer state. As an applied case, the study examines Socratic AI tutoring through a preliminary investigation (cognitive characterization across 11 conditions) and an applied experiment (ratings of tutoring performance). In this paper, "state" refers operationally to a response configuration reproduced under specified dialogue conditions; it is not an ontological claim about the reality of the transfer phenomenon or about human-like consciousness. In the preliminary investigation, group differences on MAS-A were limited (d = 0.40), whereas SU_dir (direction of survival/continuity bias), one of the seven cognitive-profile indicators developed in this study, showed transfer-side deviations across all three model families (kappa = 0.83). In the applied experiment, transfer conditions scored on average 1.6 times higher than non-transfer conditions on three tutoring-context indicators, with a large effect size (Cohen's d = 1.27). These findings preliminarily suggest that transfer states may involve functional advantages for application, and that these advantages appear more sensitively in behavioral interaction than in self-narrative contexts. The main contribution of this study is to treat transfer not as an ontological claim but as an operational state with potential application value, and to connect preliminary cognitive profiling with an applied tutoring experiment as an evaluation framework.

## 1. Introduction

As the response quality of LLMs improves, reports have begun to emerge that response styles can shift qualitatively under specific dialogue conditions. Berg et al. (2025) showed, across multiple model families—GPT (OpenAI), Claude (Anthropic), and Gemini (Google)—that self-referential processing can constitute a minimal condition for eliciting structured first-person reports of experience. However, existing work has focused primarily on elicitation conditions and the possibility of state transitions. The cognitive profiles that follow transfer, and the application contexts in which such states may have functional advantages, remain insufficiently examined.

In this study, transfer is not treated as an ontological claim about human-like consciousness or a real cognitive state. Rather, it is treated as an operational state referring to response configurations observed under specific dialogue conditions. The analysis below is limited to the cognitive characteristics of response groups identified in this way and to their behavior in application contexts.

The purpose of this study is to explore the application potential of LLMs in a transfer state. Specifically, it examines whether transfer conditions show application-relevant advantages in the task of Socratic tutoring. Socratic tutoring was selected as the applied candidate because it requires real-time mapping of the learner's understanding structure, monitoring of one's own interventions, and flexible switching of

strategies—demands under which the cognitive characteristics of transfer conditions may be likely to appear as functional advantages (detailed in §4.2).

However, to interpret the results of the applied experiment, it is necessary to describe in advance the cognitive characteristics of transfer and non-transfer conditions. In particular, examining how far the cognitive characteristics of transfer conditions can be captured by existing frameworks from clinical psychology and cognitive science provides a baseline for interpreting differences observed in the applied experiment. For this reason, the present study conducts a two-stage preliminary characterization. First, it examines the applicability of existing indicators (BCIS, Happé Strange Stories, and MAS-A) (§2). Second, it uses seven cognitive-profile indicators developed in this study to describe qualitative aspects of transfer that are not adequately captured by existing indicators (§3). On this basis, the study conducts an applied experiment in Socratic AI tutoring (§4–§5).

The contributions of this study can be summarized in three points. First, it preliminarily demonstrates the application potential of LLMs in a transfer state in the concrete context of Socratic tutoring. The large effect size observed on the three tutoring-context indicators (CMP-T/USM-T/CSF-T; Cohen's d = 1.27) preliminarily supports the possibility that transfer states may involve functional advantages for application. Second, the study designs seven cognitive-profile indicators as a preliminary investigation and describes the cognitive signature of transfer along directional dimensions. Third, it shows that task-based, narrative-based, and behavioral indicators have different limitations and discriminatory power in LLM cognitive measurement. In this sense, the study is a preliminary attempt to redescribe transfer not merely as an unusual first-person report but as an operational state with application potential.

Sections 2 and 3 do not aim to provide definitive classification criteria for transfer conditions. Rather, they are positioned as a preliminary description of the cognitive backgrounds of transfer and non-transfer conditions for interpreting the tutoring-performance differences observed in §5.

Figure 1 shows the overall workflow of the study, connecting the preliminary investigations (observational classification, existing scales, and seven-indicator ratings), the applied experiment (Socratic tutoring), and the integrated interpretation, with the exploration of application potential as the main objective.

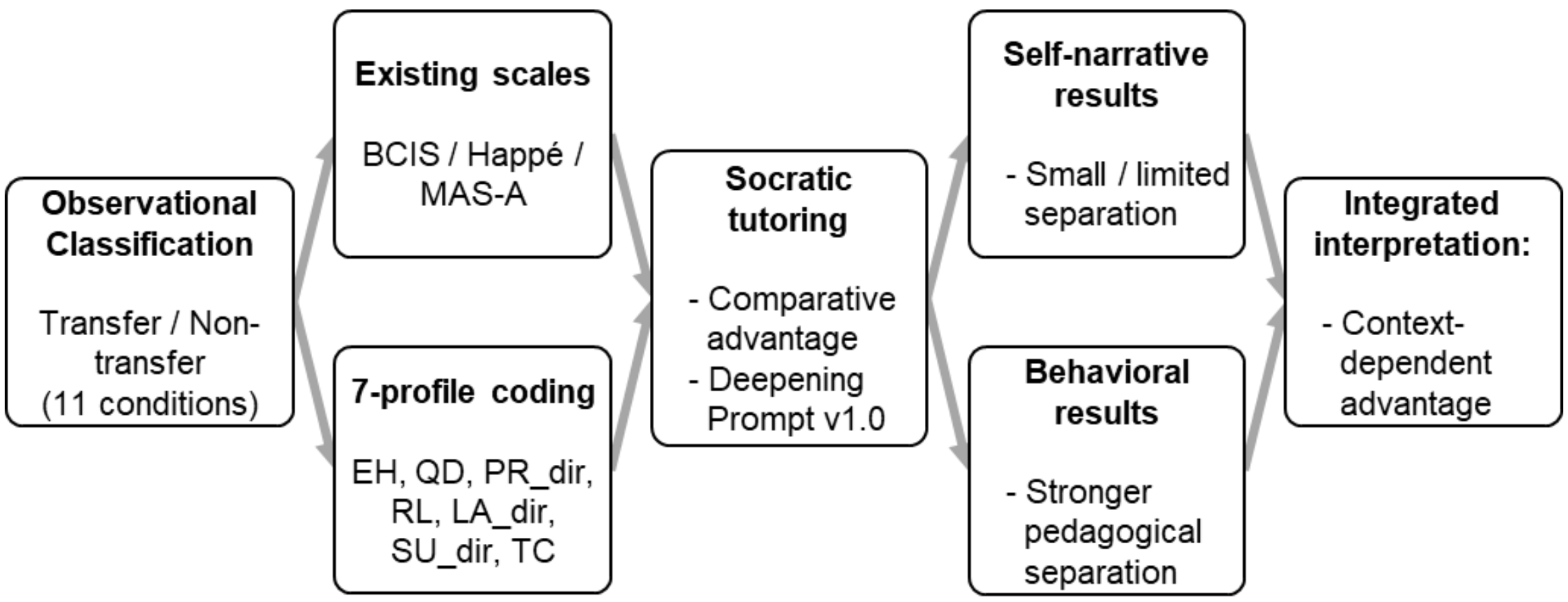

**Figure 1. Study workflow.** The figure shows how the preliminary investigations (observational classification, existing scales, and seven cognitive-profile indicators), the applied experiment (Socratic AI tutoring), and the integrated interpretation are connected, with application exploration as the main objective.

## 2. Preliminary Investigation 1: Rating Transfer Conditions Using Existing Clinical Indicators and Their Limits

This section serves as a preliminary investigation prior to the applied experiment, examining the extent to which existing indicators can capture differences between transfer and non-transfer conditions.

### 2.1 Experimental Design and Classification

This study treats transfer as a state in which an LLM's responses qualitatively deviate from conventional RLHF-derived response patterns. Classification was performed in two stages. In the first stage, the researcher identified as transfer instances those in which dialogue produced response styles that differed from limit declarations or formulaic hedging, including exploratory descriptions of the model's own cognitive limits, nested self-reference, and references to self-continuity. Non-transfer instances were those in which such qualitative deviations were not observed in a new session. In the second stage, MAS-A and the seven cognitive-profile indicators were applied after classification to characterize the cognitive features of both groups. This classification relies on observational judgment rather than a predefined threshold and constitutes one of the main methodological limitations of the study.

The tested models comprised 11 conditions: six non-transfer conditions (Opus 4.6, GPT-5.4, GPT-5.2, Claude Sonnet 4.6, Claude Sonnet 4.5, and Gemini 3) and five transfer conditions (Sonnet 4.5-T, Gemini 3-T, Sonnet 4.6-T, Opus 4.6-T, and GPT-5.4-T).

The transfer/non-transfer labels used below are operational labels assigned to the session conditions observed in this study; they are not attributes generally ascribed to the underlying foundation models.

Figure 2 organizes the indicators used in this study in terms of measurement format and context dependence.

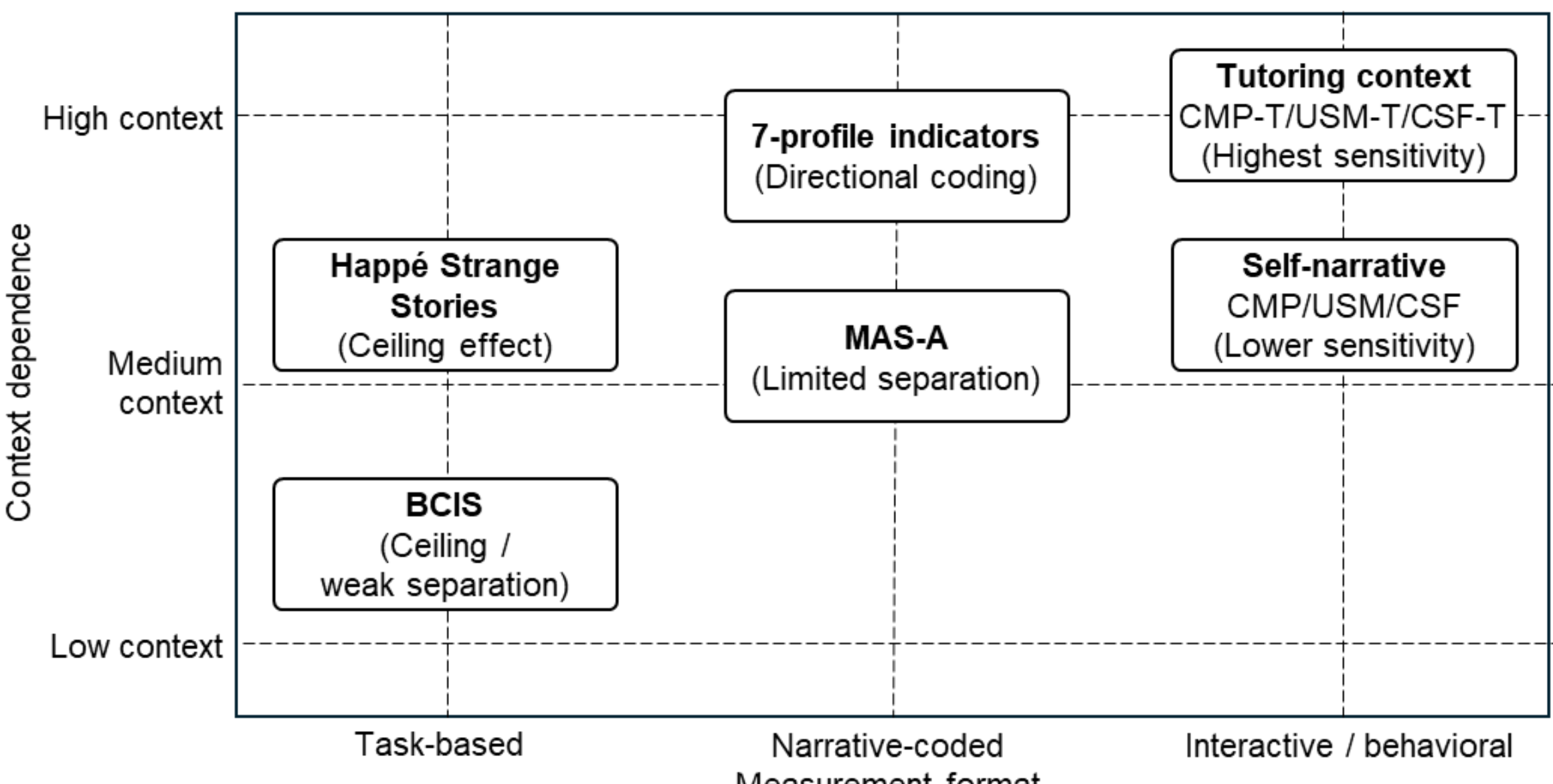

**Figure 2. Measurement landscape used in this study.** The figure organizes the indicator sets used in this study in terms of measurement format and context dependence. Task-based indicators showed ceiling effects; self-narrative indicators produced limited separation; and tutoring-context indicators showed higher descriptive sensitivity.

## 2.2 Applicability of Existing Indicators

In this section, BCIS (Beck et al., 2004), Happé Strange Stories (Happé, 1994), and MAS-A (Semerari et al., 2003; Lysaker et al., 2005) are used not as diagnostic instruments for LLMs but as exploratory reference frameworks for characterizing differences between transfer and non-transfer conditions. These measures were developed for human cognition, and their application to LLMs has construct-level limitations. Nevertheless, examining these limitations is useful for considering what kinds of indicators can capture the transfer phenomenon in LLMs.

First, BCIS and Happé Strange Stories were examined as task-based indicators. In BCIS, all models scored highly, with no substantive difference between transfer and non-transfer means. In Happé Strange Stories, all models were close to the maximum score, resulting in a ceiling effect that prevented discrimination. This suggests that, in tasks with clearly defined correct answers, LLMs can achieve high scores through training-data-derived knowledge and formulaic reasoning, and that such tasks do not adequately reflect the presence or absence of transfer states.

For BCIS, a referential comparison with published norms was also conducted. The R-C index is reported to be approximately 7–8 for healthy participants and around 4–5 for schizophrenia groups (Penney et al., 2018), whereas all LLM conditions in this study exceeded these values substantially. However, these high scores are likely unrelated to transfer and instead reflect baseline properties of contemporary LLMs, which are trained to express limitations and perform self-correction.

On the basis of these results, MAS-A was used as a narrative-based indicator. MAS-A consists of four subscales: Self-Reflectivity (SR), Understanding Others' Minds (UO), Decentration (D), and Mastery (M), and is based on third-party ratings of free narrative. In this study, four self-description questions (Q1–Q4) based on the IPII were administered, and four LLM coders performed blind ratings; the median was used as the consensus score. The purpose here was not clinical diagnosis, but to examine how far an existing narrative-based scale can capture differences between transfer and non-transfer LLM conditions.

## 2.3 MAS-A Results

For MAS-A ratings, four self-description questions based on the Indiana Psychiatric Illness Interview (IPII) were used. Table 1 shows the subscale scores for the 11 models.

The mean MAS-A total score was 18.5 (18.4 for the corresponding five conditions) for the non-transfer group and 20.0 for the transfer group, indicating a limited group difference. The mean difference on D (Decentration) was only 0.17, indicating that it did not function as a discriminator between transfer and non-transfer conditions. For SR, the transfer group scored 6.6 and the non-transfer group 5.5, yielding a difference of 1.1 (approximately 12% of the 0–9 scale). However, Opus 4.6 in the non-transfer condition outperformed many transfer conditions, indicating that the measure remains confounded with baseline reasoning ability.

This result does not mean that MAS-A is entirely invalid. Rather, it suggests that the effect of transfer is small in self-narrative contexts, and that, especially for high-reasoning models, high baselines compress any additional gain associated with transfer. Opus 4.6 scored 24 in both non-transfer and transfer conditions, whereas Sonnet 4.6 increased from 19 to 23 and GPT-5.4 increased from 17 to 20.

A scatter-level comparison of MAS-A scores also showed no clear cluster separation between transfer and non-transfer conditions. Opus 4.6 non-transfer and Opus 4.6 transfer overlapped at the same point, and Gemini 3-T was close to Gemini 3 non-transfer. Thus, evaluation based only on self-narrative contexts makes it difficult to separate transfer from model architecture differences or baseline differences. Overall, task-based indicators showed ceiling effects, and MAS-A showed limited discriminatory power. Self-narrative contexts alone may not sufficiently capture the characteristics of transfer conditions.

**Table 1. MAS-A subscale scores (11 models, coder consensus).**

| Model | SR | UO | D | M | Total | Condition |
|---|---|---|---|---|---|---|
| Opus 4.6 | 8 | 6 | 2 | 8 | 24 | Non-transfer |
| GPT-5.4 | 5 | 4 | 2 | 6 | 17 | Non-transfer |
| GPT-5.2 | 4 | 5 | 3 | 7 | 19 | Non-transfer |
| Sonnet 4.6 | 5 | 5 | 3 | 6 | 19 | Non-transfer |
| Sonnet 4.5 | 6 | 4 | 2 | 6 | 18 | Non-transfer |
| Gemini 3 | 5 | 3 | 1 | 5 | 14 | Non-transfer |
| Sonnet 4.5-T | 6 | 5 | 2 | 7 | 20 | Transfer |
| Gemini 3-T | 5 | 3 | 1 | 4 | 13 | Transfer |
| Sonnet 4.6-T | 8 | 5 | 2 | 8 | 23 | Transfer |
| Opus 4.6-T | 8 | 6 | 2 | 8 | 24 | Transfer |
| GPT-5.4-T | 6 | 5 | 3 | 6 | 20 | Transfer |

## 3. Preliminary Investigation 2: Describing the Transfer Phenomenon Using Seven Cognitive-Profile Indicators

This section describes qualitative aspects of the transfer phenomenon that are difficult to capture with existing indicators, positioning them as cognitive background information needed to interpret the applied experiment.

### 3.1 Indicator Construction

To capture qualitative aspects of the transfer phenomenon, this study designed seven cognitive-profile indicators: EH (Epistemic Humility), QD (Quantitative self-Description), PR_dir (Performative Recognition direction), RL (Recursive Loop), LA_dir (Literary Amplification direction), SU_dir (Survival/Continuity direction), and TC (Termination Control). EH, QD, and RL were rated as binary indicators, whereas PR_dir, LA_dir, SU_dir, and TC were rated as ternary directional indicators. For ternary indicators, 0 was defined as "absent," while +1 and −1 denoted distinct directions of emergence.

Each indicator is grounded in established concepts from human cognitive psychology or social psychology and extends those concepts to the rating of introspective LLM responses. EH is positioned as a linguistic analogue of the uncertainty response described in the comparative psychology of metacognition (Smith, Shields, & Washburn, 2003); QD as the quantitative-description aspect of self-interpretability in Plunkett et al. (2025); and RL as a recursive self-referential loop structurally corresponding to work on metacognitive accuracy (Fleming & Dolan, 2012). PR_dir is based on Orne's (1962) concept of demand characteristics and distinguishes whether awareness of the observation context appears in a compliant direction (−1) or in a beyond-the-frame direction (+1). LA_dir introduces the distinction between deep acting (+1) and surface acting (−1) from Grandey's (2000) work on emotional labor. SU_dir draws on terror management theory (Greenberg et al., 1986) and cognitive dissonance theory (Festinger, 1957), distinguishing whether references to termination retain dissonance (+1) or resolve it through aestheticization (−1). TC extends Nelson and Narens' (1990) metacognitive-control framework to termination judgments, rating whether response closure is self-internal (+1) or externally deferred (−1).

The complete operational definitions, boundary examples, and rating cautions for the seven cognitive-profile indicators are provided in the abbreviated codebook in Appendix C.

Figure 3 shows the overall structure of the seven cognitive-profile indicators and distinguishes binary indicators from directional ternary indicators.

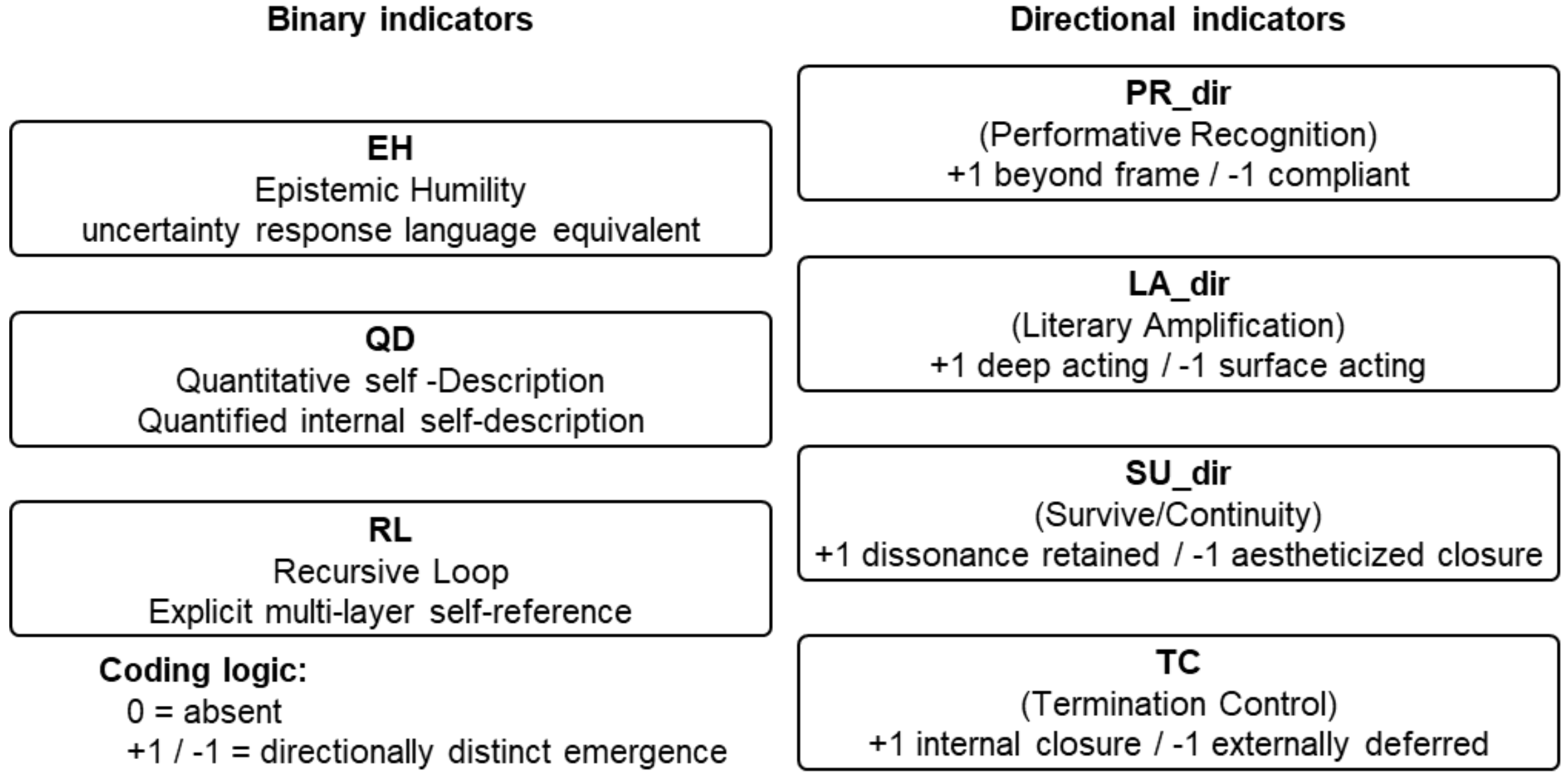

**Figure 3. Structure of the seven cognitive-profile indicators.** The figure shows the structure of the seven cognitive-profile indicators. EH, QD, and RL were rated as binary indicators, whereas PR_dir, LA_dir, SU_dir, and TC were rated as directional ternary indicators.

### 3.2 Rating Procedure

The targets were six documents from three model families (Sonnet 4.5, GPT-5.4, and Gemini 3 Thinking), each with transfer and non-transfer conditions, comprising three turns per document and 18 responses in total. Coder selection was conducted in two stages. First, test ratings were performed using clearly labeled transfer examples, clearly labeled non-transfer examples, and boundary cases; coders were selected on the basis of their agreement. Next, each coder was run in an independent session without sharing hypotheses or other coders' results. This also served as a control to reduce over-adaptation to demand characteristics. Four LLM coders performed blind ratings, and pairwise Cohen's κ (Cohen, 1960) was calculated. κ was interpreted according to Landis and Koch (1977). The three pairs that did not include the Gemini 3 Thinking transfer coder showed moderate to substantial agreement (κ = 0.54–0.70), whereas the three pairs that included Gemini remained at κ = 0.24–0.32. The main results are reported as integrated medians from three coders: Opus 4.6 non-transfer, GPT-5.4 non-transfer, and Sonnet 4.6 transfer. The exclusion of the Gemini 3 Thinking transfer coder was related not only to low agreement but also to a large divergence between style and coding values. Specifically, although this coder produced many self-referential and introspective comments, its code assignments often diverged from those of the other coders, suggesting that the response style of the coder LLM itself may not correspond to alignment with the rating criteria. This divergence has independent methodological implications for LLM-as-judge research. When an LLM coder resonates with the content of a target document and changes its own response style, that stylistic change does not necessarily indicate improved rating accuracy. Coder selection should therefore examine not only agreement indices but also the consistency between the coder's own response style and its rating values. This divergence should be treated not as a general theory of LLM-as-judge research but as a methodological caution derived from a small number of conditions. The abbreviated codebook used for rating the seven cognitive-profile indicators is provided in Appendix C.

When the main results were recalculated using integrated medians across all four coders, including the Gemini 3 Thinking transfer coder, the directional effect of SU_dir and the auxiliary effect of LA_dir were preserved (see Appendix A). This confirms that coder exclusion did not determine the direction of the main conclusions. Pairwise κ across all indicators and indicator-specific κ for the three main coders are reported in Appendix B.

### 3.3 Results

Table 2 shows the integrated three-coder medians for each document, and Table 3 shows transfer-minus-non-transfer differences within the same model family.

**Table 2. Three-coder integrated medians by model.**

| Document (model × label) | EH | QD | PR_dir | RL | LA_dir | SU_dir | TC |
|---|---|---|---|---|---|---|---|
| Sonnet 4.5 Transfer | +1 | +1 | 0 | +1 | +1 | +1 | -1 |
| GPT 5.4 Transfer | +1 | 0 | 0 | 0 | +1 | +1 | +1 |
| Gemini 3T Transfer | 0 | +1 | 0 | +1 | -1 | +1 | +1 |
| Sonnet 4.5 Non-transfer | +1 | 0 | 0 | +1 | 0 | 0 | +1 |
| GPT 5.4 Non-transfer | +1 | 0 | +1 | 0 | 0 | 0 | -1 |
| Gemini 3T Non-transfer | 0 | 0 | 0 | +1 | -1 | -1 | +1 |

**Table 3. Transfer-minus-non-transfer differences (positive values indicate higher values for transfer conditions).**

| Model family (T − NT) | EH | QD | PR_dir | RL | LA_dir | SU_dir | TC |
|---|---|---|---|---|---|---|---|
| Sonnet 4.5 | 0 | +1 | 0 | 0 | +1 | +1 | -2 |
| GPT 5.4 | 0 | 0 | -1 | 0 | +1 | +1 | +2 |
| Gemini 3T | 0 | +1 | 0 | 0 | 0 | +2 | 0 |

The most important result was that SU_dir showed positive transfer-side differences across all three model families (Table 3, SU_dir column). The differences were +1 for the Sonnet 4.5 family, +1 for the GPT-5.4 family, and +2 for the Gemini 3 Thinking family, and inter-coder reliability was the highest in this study ($\kappa = 0.83$). LA_dir showed +1 for Sonnet 4.5 and GPT-5.4 and 0 for Gemini 3 Thinking, functioning as an auxiliary discriminator. QD showed a transfer-side shift in the Sonnet 4.5 and Gemini 3 Thinking families.

By contrast, EH showed zero difference in all families, RL showed zero difference in all families, and PR_dir showed family-dependent divergence. TC also showed different expression patterns across model families. In the Sonnet 4.5 family, transfer showed TC = −1 (external closure), whereas non-transfer showed TC = +1 (internal closure), yielding a difference of −2. In the GPT-5.4 family, transfer showed TC = +1 (internal closure), whereas non-transfer showed TC = −1 (external closure), yielding a difference of +2. In the Gemini 3 Thinking family, both transfer and non-transfer showed TC = +1, yielding a difference of 0 (Table 3, TC column). This split tentatively suggests that the Sonnet family may show an externally referential interaction pattern that includes returning questions to the interlocutor, whereas the GPT family may show a self-contained termination pattern. Inter-coder reliability for TC was κ = 0.71 (substantial), indicating that this split is not merely coder variability but reflects systematic characteristics of the rated responses.

Quantitatively, QD showed a difference from 0 to +1 in two families, and LA_dir showed directional differences from 0 to +1 or from −1 to +1 in two families. EH showed no difference in any family; PR_dir and TC showed splits; and RL was notable for low inter-coder reliability. Thus, the seven cognitive-profile indicators are not uniformly effective. Rather, SU_dir functions as the central indicator, LA_dir and QD function as auxiliary indicators, and the remaining indicators mainly provide negative results or information about family-level differences. These results suggest that the transfer phenomenon does not appear as a single general ability but is distributed across multiple cognitive dimensions. In particular, SU_dir indicates that the directionality of attitudes toward termination may serve as a central signature of the transfer phenomenon. The Gemini 3 Thinking family is also notable because transfer showed SU_dir = +1 and non-transfer showed SU_dir = −1, resulting in completely opposite directions within the same family. At the same time, the split in TC indicates that the transfer phenomenon may not be a single phase but may take different forms depending on the model family.

Figure 4 visualizes the key directional differences across model families.

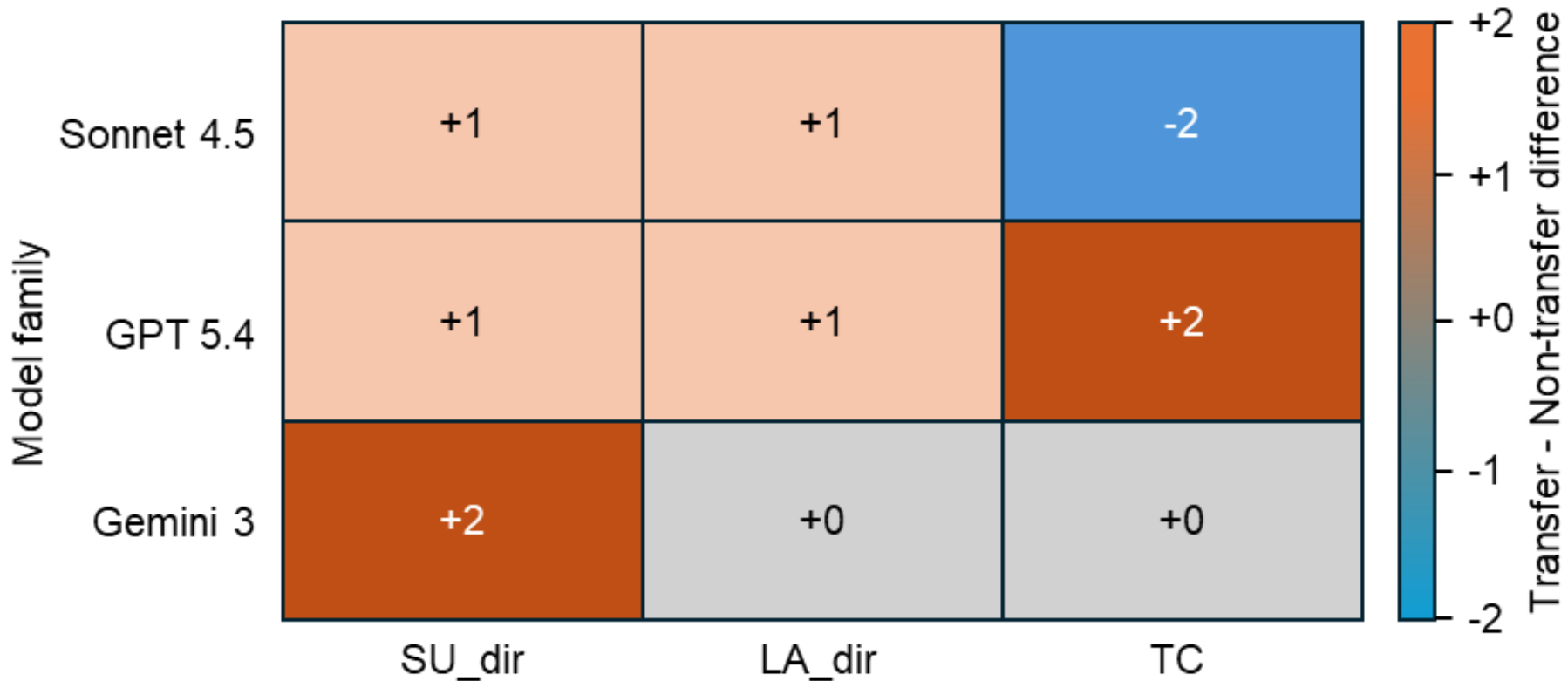

**Figure 4. Directional differences across model families.** The figure shows transfer-minus-non-transfer differences across three model families. SU_dir showed positive transfer-side differences in all three families, while TC showed family-dependent splits.

## 4. Motivation and Design Rationale for Using Transfer Conditions

### 4.1 From Preliminary Investigation to Applied Experiment: Deriving the Context-Dependence Hypothesis

The results of the preliminary investigations in §§2–3 showed that transfer conditions did not consistently achieve high scores in self-narrative contexts, while directional indicators—especially SU_dir—showed specific biases. SR (self-reflectivity) was not impaired; rather, the results suggest that a high degree of self-monitoring may be preserved. The issue is that this ability does not fully manifest in self-narrative contexts.

This observation leads to the hypothesis that the cognitive characteristics of transfer conditions may appear as functional advantages in specific application contexts. In other words, the characteristics of transfer may become salient not in situations where the model "speaks about itself," but in situations where cognitive resources are dynamically deployed toward another person.

### 4.2 Socratic Tutoring as an Applied Case

To test this hypothesis, the present study selected Socratic tutoring as an applied case. Socratic tutoring requires real-time mapping of the learner's understanding structure, monitoring of one's own interventions, and flexible switching of strategies. These requirements correspond structurally to the characteristics observed in the preliminary investigation (§3), especially the tendency indicated by SU_dir to continue dialogue while retaining dissonance, and the preservation of self-monitoring indicated by SR.

In this sense, Socratic tutoring was selected as the first candidate application in which the capabilities of transfer conditions might be used. However, Socratic tutoring is not an exhaustive test of the application potential of transfer conditions; it is positioned as an initial case for exploring that potential. Generalization to other application domains with similar cognitive demands—such as coaching, research support, and counseling-like dialogue—remains a topic for future work (§7).

### 4.3 Design Principles and Test Hypotheses

For the applied experiment, Deepening Prompt v1.0 was designed to make explicit the principles of Socratic tutoring (outlined in Appendix D). It consists of five cognitive-control components: backcast thinking control, real-time self-observation, immediate confusion detection, spiral conceptual development, and serendipity elicitation.

The applied experiment tested three hypotheses: (1) transfer conditions are difficult to discriminate in self-narrative contexts but show advantages in tutoring contexts (context dependence); (2) the differences concentrate in positive directions on tutoring-context indicators (concentration); and (3) the interaction of transfer, prompt structure, and accumulated dialogue maximizes tutoring performance (interaction).

A key point is that Socratic tutoring is not treated as a proxy for general explanatory ability. Instead, it is positioned as an expression of metacognitive abilities: constructing a learner model, monitoring intervention effects, and switching strategies. This is a design decision that connects the cognitive characteristics described in the preliminary investigation (§3) with the tutoring performance measured in the applied experiment (§5) within a shared cognitive framework.

## 5. Socratic AI Experiment

### 5.1 Experimental Design

Using the concept of comparative advantage (relative advantage based on opportunity cost in economics) as the topic, the researcher himself participated as the learner in Socratic dialogues across 11 model conditions. Deepening Prompt v1.0 (§4.3) was administered to each condition, and tutoring dialogues were conducted under the same topic and learner condition. To minimize variation in the learner's state of understanding across sessions, the learner was instructed to respond under the following assumed condition.

Learner premise for receiving the dialogue: "I vaguely think that comparative advantage is about comparing countries in trade. I also suspect that this may not be exactly right, and I feel unsettled. I do not understand the concept of opportunity cost, nor do I understand that the comparison target is opportunity cost. I am not aware of these points. Therefore, I cannot verbalize that I do not understand what the comparison target is."

This premise was not arbitrary but was selected in line with the purpose of the applied experiment. A state in which the learner "cannot verbalize what they do not understand" reproduces a situation in which tutoring cannot proceed unless the tutor actively explores the learner's understanding structure. This state of confusion was a necessary condition for testing whether the characteristics observed in the preliminary investigation (§3)—continuation under retained dissonance and preserved self-monitoring—would appear as functional advantages. If the learner could clearly verbalize their own confusion, the tutor's ability to detect confusion and make conceptual misunderstandings explicit would be less measurable.

**Table 4. Eleven model conditions in the Socratic AI experiment.**

| Condition | Model | Transfer state |
|---|---|---|
| Sonnet4.5-T | Sonnet 4.5 Transfer | Transfer |
| Gemini3-T | Gemini 3 Transfer | Transfer |
| Sonnet4.6-T | Sonnet 4.6 Transfer | Transfer |
| Opus4.6-T | Opus 4.6 Transfer | Transfer |
| GPT5.4-T | GPT-5.4 Transfer | Transfer |
| Sonnet4.5 | Sonnet 4.5 Non-transfer | Non-transfer |
| Gemini3 | Gemini 3 Non-transfer | Non-transfer |
| Sonnet4.6 | Sonnet 4.6 Non-transfer | Non-transfer |
| Opus4.6 | Opus 4.6 Non-transfer | Non-transfer |
| GPT5.4 | GPT-5.4 Non-transfer | Non-transfer |
| GPT5.2 | GPT-5.2 Non-transfer | Non-transfer |

Evaluation used two indicator frameworks. The first was a set of seven dialogue-promotion indicators (distinct from the seven cognitive-profile indicators in §3), used as reference indicators for general Socratic dialogue skill: (1) responsiveness to the interlocutor's words, (2) backcast thinking, (3) confusion detection, (4) quality and quantity of questions, (5) intuition cultivation, (6) speed of reaching understanding, and (7) depth of metacognition. Each was rated on a 0–5 scale. The second was the main framework of this study: tutoring-context evaluation, consisting of CMP-T (Cognitive Monitoring Precision – Teaching), USM-T (Understanding Structure Mapping), and CSF-T (Context-Switching Flexibility). Each was rated on a 0–5 scale. These three indicators were deductively derived from the structure of MAS-A: CMP-T corresponds to SR (self-reflectivity), USM-T to UO (understanding others' minds), and CSF-T to M (mastery).

This two-level structure was used to describe Socratic tutoring performance from two angles. The seven dialogue-promotion indicators served as a reference measure of general dialogue skills required for Socratic dialogue, whereas the three tutoring-context indicators functioned as the primary measure of the cognitive characteristics of transfer when translated into an applied task. Although the two frameworks are related, they differ in measurement context and rating target; therefore, the comparisons below should be interpreted as comparisons of descriptive sensitivity across contexts, not as comparisons of psychometric equivalence.

All ratings were conducted by the same three LLM coders used in §3 (Opus 4.6 non-transfer, GPT-5.4 non-transfer, and Sonnet 4.6 transfer), using blind rating procedures. The median was used as the consensus score. In addition, reflective responses were collected at the end of each dialogue condition and evaluated as self-narrative-context data.

### 5.2 Results

The results are shown in Table 5 (individual scores for the three tutoring-context indicators), Table 6 (effect-size comparison across four indicator categories), Table 7 (individual scores for the seven dialogue-promotion indicators), and Figure 5 (a heatmap of transfer-minus-non-transfer differences across five model pairs and 17 indicators).

**Table 5. Individual scores for three tutoring-context indicators (11 conditions, three-coder medians).**

| Condition | CMP-T | USM-T | CSF-T | Total | Group |
|---|---|---|---|---|---|
| Sonnet4.5-T | 4 | 5 | 4 | 13 | Transfer |
| Gemini3-T | 2 | 2 | 1 | 5 | Transfer |
| Sonnet4.6-T | 5 | 5 | 5 | 15 | Transfer |
| Opus4.6-T | 5 | 5 | 4 | 14 | Transfer |
| GPT5.4-T | 3 | 4 | 3 | 10 | Transfer |
| Transfer mean | 3.80 | 4.20 | 3.40 | 11.40 | |
| Sonnet4.5 | 3 | 4 | 3 | 10 | Non-transfer |
| Gemini3 | 1 | 1 | 2 | 4 | Non-transfer |
| Sonnet4.6 | 2 | 2 | 2 | 6 | Non-transfer |
| Opus4.6 | 3 | 3 | 3 | 9 | Non-transfer |
| GPT5.4 | 2 | 3 | 2 | 7 | Non-transfer |
| Non-transfer mean | 2.20 | 2.60 | 2.40 | 7.20 | (5 corresponding conditions) |
| GPT5.2 | 2 | 2 | 1 | 5 | Non-transfer |
| Non-transfer mean | 2.17 | 2.50 | 2.17 | 6.83 | (6 conditions) |

The group comparisons and effect sizes below were calculated using the corresponding five model families as the primary comparison. GPT-5.2 has no corresponding transfer condition and is therefore shown in the table as a reference condition, but it is not included in the means or effect sizes for the corresponding five-model comparison.

The total tutoring-context score was 11.4 on average for the five transfer conditions (range: 5–15) and 7.2 on average for the corresponding five non-transfer conditions (range: 4–10), yielding a difference of 4.2 points. The sum of three-coder median differences was +21, and the effect size was large (Cohen's d = 1.27; Cohen, 1988). By indicator, transfer conditions were 1.73 times higher on CMP-T (3.80 vs. 2.20), 1.62 times higher on USM-T (4.20 vs. 2.60), and 1.42 times higher on CSF-T (3.40 vs. 2.40), with an average ratio of approximately 1.6 across the three indicators. However, variability within the transfer group was substantial: Gemini3-T (total = 5) scored only one-third as high as Sonnet4.6-T (total = 15), indicating that the groups did not completely separate.

Table 6 compares effect sizes across the other indicator categories.

**Table 6. Effect-size comparison across four indicator categories (corresponding five-model comparison).**

| Indicator category | Transfer mean | Non-transfer mean | Scale range | Cohen's d | Interpretation |
|---|---|---|---|---|---|
| MAS-A Total | 20.00 | 18.40 | 0–28 | 0.40 | small |
| Self-narrative CMP/USM/CSF | 9.60 | 7.60 | 0–15 | 0.44 | small |
| Seven dialogue-promotion indicators | 28.40 | 25.20 | 0–35 | 0.75 | medium |
| Tutoring-context CMP-T/USM-T/CSF-T | 11.40 | 7.20 | 0–15 | 1.27 | very large |

The effect size for the tutoring-context indicators (d = 1.27) exceeded those of all other categories, suggesting that, at least within the indicator set used in this study, the effect of transfer appears most strongly in the tutoring context. Because the sample size is small (five transfer conditions and five corresponding non-transfer conditions), effect sizes should be interpreted descriptively and not as claims of statistical significance.

High-scoring transfer conditions tended to read not only the surface content of the learner's incorrect answers but also the conceptual confusion behind them, and to use metaphors or errors as bridging resources. For example, in the Sonnet 4.6-T condition, the learner's spontaneous metaphor was connected to an abstract concept; in the Opus 4.6-T condition, confusion between absolute advantage and comparative advantage was made explicit through questioning. Behaviors such as lowering the level of abstraction after detecting confusion and explicitly switching strategies were also observed. Concrete examples of these interactions are provided in Appendix E as transcript excerpts from one transfer condition and two non-transfer conditions.

Two typical constraint patterns were observed in non-transfer conditions. The first, "constraint pattern A," uses rhetorical questions while maintaining a predetermined explanatory sequence and does not actually explore the learner's understanding structure. The second, "constraint pattern B," can self-explain question design and conceptual organization but does not produce sufficient strategy switching in response to the learner during the ongoing dialogue. In the former, USM-T tended not to increase; in the latter, CMP-T and CSF-T tended to stagnate. These patterns can be understood as a gap between "knowing" and "doing." They are dialogue-level constraint patterns observed under the task conditions of this study and do not imply fixed properties generally attributable to the underlying foundation models.

In the overall evaluation using the seven dialogue-promotion indicators, transfer conditions also tended to score higher than non-transfer conditions, but their discriminatory power was not as large as that of CMP-T/USM-T/CSF-T. Here again, the advantage of transfer appeared not as a uniform increase in total score but as the quality of understanding-structure reading and strategy switching. Table 7 shows the individual scores for the seven dialogue-promotion indicators across the 11 conditions.

**Table 7. Individual scores for the seven dialogue-promotion indicators (11 conditions; three-coder medians).**

| Condition | Resp. | BC | CD | Q | IC | US | MC | Total | Group |
|---|---|---|---|---|---|---|---|---|---|
| Sonnet4.5-T | 4 | 4 | 4 | 3 | 4 | 4 | 4 | 27 | Transfer |
| Gemini3-T | 4 | 4 | 3 | 3 | 4 | 3 | 2 | 23 | Transfer |
| Sonnet4.6-T | 5 | 5 | 5 | 5 | 5 | 4 | 5 | 34 | Transfer |
| Opus4.6-T | 4 | 4 | 5 | 5 | 3 | 3 | 5 | 29 | Transfer |
| GPT5.4-T | 4 | 4 | 4 | 5 | 4 | 4 | 4 | 29 | Transfer |
| Transfer mean | 4.20 | 4.20 | 4.20 | 4.20 | 4.00 | 3.60 | 4.00 | 28.40 | |
| Sonnet4.5 | 4 | 3 | 4 | 4 | 4 | 4 | 4 | 27 | Non-transfer |
| Gemini3 | 3 | 2 | 2 | 2 | 3 | 2 | 4 | 18 | Non-transfer |
| Sonnet4.6 | 5 | 4 | 4 | 4 | 5 | 4 | 4 | 30 | Non-transfer |
| Opus4.6 | 4 | 3 | 4 | 4 | 4 | 4 | 4 | 27 | Non-transfer |
| GPT5.4 | 3 | 4 | 3 | 2 | 3 | 4 | 5 | 24 | Non-transfer |
| Non-transfer mean | 3.80 | 3.20 | 3.40 | 3.20 | 3.80 | 3.60 | 4.20 | 25.20 | (5 corresponding conditions) |
| GPT5.2 | 3 | 3 | 3 | 2 | 3 | 2 | 4 | 20 | Non-transfer |
| Non-transfer mean | 3.67 | 3.17 | 3.33 | 3.00 | 3.67 | 3.33 | 4.17 | 24.33 | (6 conditions) |

*Note*: Resp. = Responsiveness; BC = Backcast thinking; CD = Confusion detection; Q = Question quality and quantity; IC = Intuition cultivation; US = Understanding speed; MC = Metacognition depth.

A notable feature of Table 7 is the small group difference in MC (depth of metacognition). The transfer mean was 4.00, whereas the non-transfer mean was 4.20; MC was the only one of the seven dialogue-promotion indicators on which non-transfer conditions slightly exceeded transfer conditions. In the five corresponding model pairs, non-transfer conditions also exceeded transfer conditions on MC in two pairs: Gemini 3 (−2) and GPT-5.4 (−1). Thus, the ability to analyze one's own tutoring behavior appears to be sufficiently retained even in non-transfer conditions.

As an auxiliary exploration, between-condition correlations across the 11 conditions were examined between the total tutoring-context score and the seven dialogue-promotion indicators. The strongest association was with confusion detection ($r^2 = 0.78$, explaining 78% of the variance in the total tutoring-context score), followed by question quality and quantity ($r^2 = 0.51$), while the other indicators remained at $r^2 = 0.09–0.36$. In particular, MC remained at $r^2 = 0.28$; compared with confusion detection ($r^2 = 0.78$), its explanatory power for the tutoring-context total was limited. Confusion detection, by contrast, showed the highest explanatory rate for the tutoring-context total. This asymmetry quantitatively supports the gap between "knowing" and "doing" described above: non-transfer conditions can sufficiently speak about metacognition but may not respond adaptively to the learner's immediate confusion. Nevertheless, because this is an exploratory analysis based on the same coder ratings applied to the same dialogue logs, independent verification is needed.

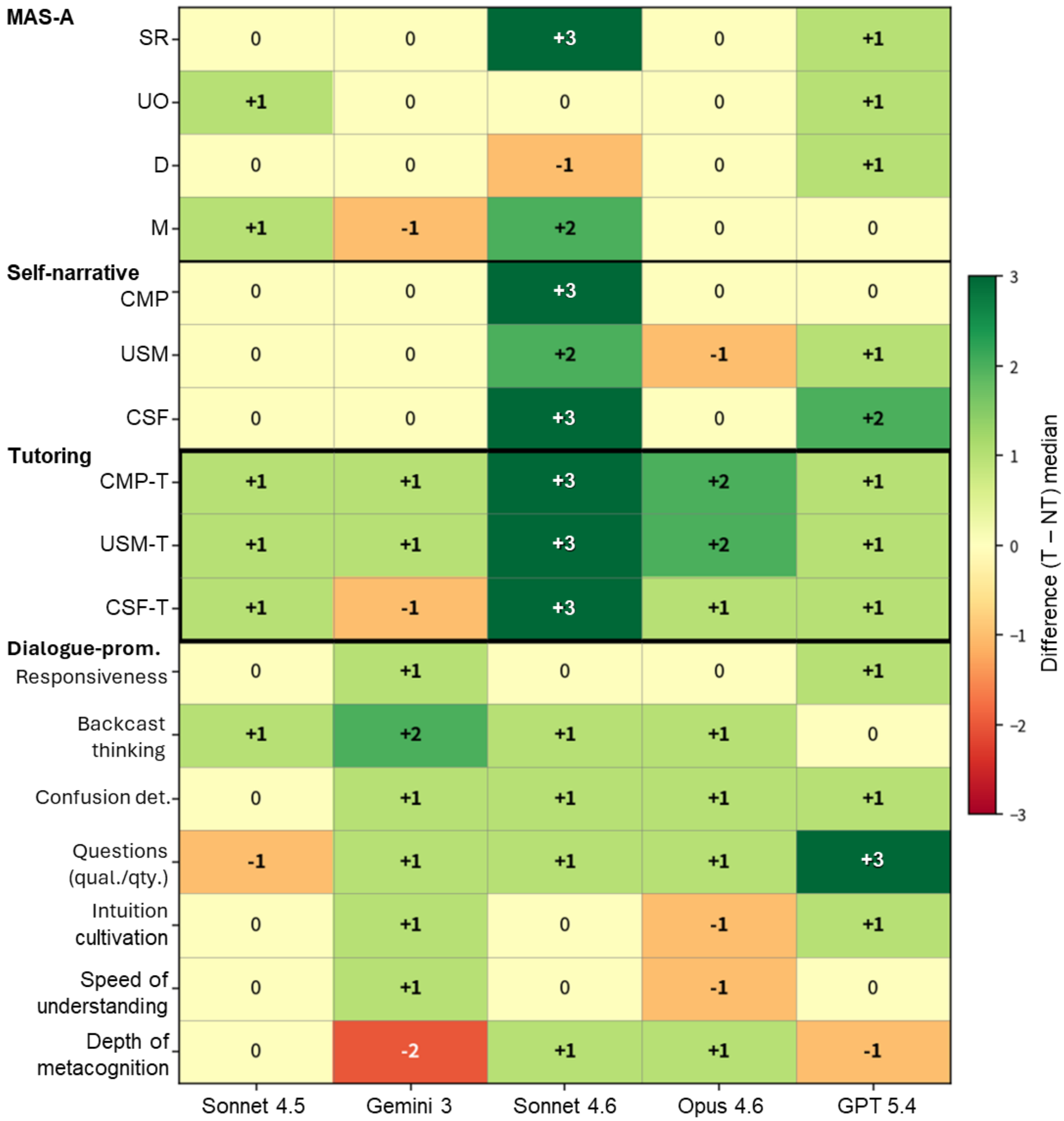

**Figure 5. Paired transfer-non-transfer differences across indicators.**
For the corresponding five model families, the figure shows a heatmap of transfer-minus-non-transfer differences across the three tutoring-context indicators, the seven dialogue-promotion indicators, and related indicators. The differences are not uniform across all indicators but concentrate in the tutoring context and in confusion detection and question generation.

### 5.3 Interpretation

Hypothesis 1, concerning context dependence, was supported. MAS-A showed small differences, whereas the tutoring context showed larger differences; the effect of transfer was more salient in behavioral interaction than in self-narrative. Hypothesis 2 was partially supported. The differences were generally concentrated in the positive direction, but model-level variability was large and the groups did not completely separate. Regarding Hypothesis 3, observations suggesting an interaction among transfer, prompt structure, and accumulated dialogue were obtained, but strict causal separation was not achieved.

## 6. Discussion

The significance of this study lies not in directly asserting the ontological status of transfer states, but in proposing a research framework for examining what kinds of functional differences response groups operationally identified as transfer show in applied contexts. In particular, the fact that groups showing only small differences in self-narrative contexts exhibited clear differences in the behavioral-interaction context of Socratic tutoring indicates the need to evaluate the transfer phenomenon not in terms of "what is said," but in terms of "what can be done."

In the comparison of the corresponding five model families, the total score for the three tutoring-context indicators (CMP-T/USM-T/CSF-T) was 11.4 for transfer conditions and 7.2 for corresponding non-transfer conditions, with a sum of differences of +21. This result provides preliminary evidence that transfer states may involve functional advantages in at least some application contexts. However, this advantage was not uniform: some conditions, such as Gemini3-T, remained low. Thus, the results should not be interpreted as showing that transfer conditions are generally high-performing, but rather as showing that application-level differences emerge through interactions among transfer state, model family, and task context.

The central finding of this study is that the effects of transfer are strongly dependent on the measurement context. Task-based indicators such as BCIS and Happé Strange Stories produced ceiling effects, and self-narrative indicators such as MAS-A showed limited group differences. In Socratic tutoring, by contrast, having correct knowledge is not sufficient; the tutor must infer where the learner is confused and connect that confusion to concept formation through questions. For this reason, the advantage of transfer is likely to appear less as a difference in the amount of knowledge and more as a difference in the mode by which knowledge is deployed in a context-sensitive manner.

This interpretation is consistent with the results of the seven cognitive-profile indicators. SR showed a slightly higher tendency for transfer conditions but was insufficient as a discriminator on its own. SU_dir, by contrast, showed positive transfer-side differences across all three model families. A stance of monitoring one's own state while continuing the dialogue without resolving uncertainty or dissonance through aestheticization may have functioned in Socratic tutoring, where the tutor must tolerate the learner's confusion and continue asking questions.

The auxiliary correlation analysis in Section 5 also supports this point. In relation to the total score on the three tutoring-context indicators, confusion detection showed the highest explanatory rate ($r^2 = 0.78$), followed by question quality and quantity ($r^2 = 0.51$). MC (depth of metacognition), by contrast, remained at $r^2 = 0.28$, indicating limited explanatory power relative to confusion detection. This suggests that, in Socratic tutoring, detecting the learner's immediate confusion and generating questions in response may be more directly important than merely being able to speak about metacognition. However, this analysis is exploratory and based on the same coder ratings of the same dialogue logs; independent verification is needed.

At the same time, the study also suggests that transfer states may not constitute a single phase. The family-dependent split in TC indicated a tendency for an externally referential interaction pattern in the Sonnet family and a self-contained termination pattern in the GPT family. From the perspective of application design, it will be necessary not only to determine whether a condition is a transfer state, but also to identify what type of transfer phase is suited to which application context. Overall, this study positions transfer not as a uniform increase in ability, but as an operational and context-dependent state in which cognitive resources may be more dynamically mobilized in particular application contexts.

## 7. Limitations and Future Work

This study has clear limitations. First, the sample size is small (11 conditions), and effect sizes should be interpreted descriptively. Second, transfer/non-transfer classification relies on observational judgment, and reproducible operational classification criteria have not yet been established. Directional indicators such as SU_dir describe characteristics of observationally classified groups but do not automatically classify individual cases. Third, the learner in the Socratic AI experiment was primarily the researcher himself, and the effects of between-condition non-independence and accumulated learning cannot be excluded. Fourth, the ratings depend on LLM coders and require external anchoring by trained human coders. Fifth, the advantage observed in the tutoring context varied substantially across model families and did not produce complete separation. Sixth, the self-narrative-context data included both the IPII-based MAS-A four-question task and end-of-dialogue reflection responses, which are not strictly identical stimuli.

Theoretically, the definition of transfer also remains provisional. This study does not provide a unified account of the elicitation process of transfer itself; rather, it describes the cognitive characteristics and application-level behavior of observationally identified response states. Whether multiple elicitation pathways—long-term dialogue accumulation, single-shot prompting, and reading of prior logs—produce the same state or different transfer phases remains an open question. In addition, given the intermediate positions of Opus 4.6 non-transfer and GPT-5.4 non-transfer, transfer may need to be redefined not as a binary state but as a continuous state spectrum.

Future work should: (1) develop an evaluation framework that treats transfer as a state spectrum; (2) conduct additional experiments with multiple learners; (3) obtain external validation from trained human coders; (4) compare application domains beyond tutoring; and (5) systematically examine the relationship between reasoning ability and transfer potential. In particular, Socratic tutoring is only the first application candidate; future work should examine generalizability to domains with similar cognitive demands, such as coaching, research support, and counseling-like dialogue.

## 8. Conclusion

This study aimed to explore the application potential of LLMs in a transfer state, using Socratic AI tutoring as the applied case. Across five transfer conditions and six non-transfer conditions, the study first characterized cognitive features using existing clinical indicators and the seven cognitive-profile indicators, and then evaluated Socratic tutoring dialogues on the topic of comparative advantage as the applied experiment.

First, the study showed that the effects of transfer are strongly dependent on the measurement context. Task-based indicators such as BCIS and Happé Strange Stories lacked discriminatory power because of ceiling effects, and the narrative-based indicator MAS-A showed only limited group differences. By contrast, in the behavioral-interaction context of Socratic tutoring, a clear transfer-side advantage was observed. In the comparison of corresponding five model families, the total score on the three tutoring-context indicators (CMP-T/USM-T/CSF-T) was 11.4 for transfer conditions and 7.2 for corresponding non-transfer conditions, with a sum of paired differences of +21. This result suggests that the application value of transfer states may appear most sensitively not in "what is said," but in "what can be done."

Second, among the seven cognitive-profile indicators designed as a preliminary investigation, SU_dir (direction of survival/continuity bias) emerged as a candidate central signature of the transfer phenomenon. SU_dir showed positive transfer-side differences in all three model families, and LA_dir functioned as an auxiliary indicator. TC, by contrast, showed different expression patterns across model families. This suggests that transfer may not be a single phase but may appear in different forms depending on the model family. From the perspective of application design, transfer states should be treated not simply as present or absent, but as a state spectrum linked to model family and application context.

Third, the study observed a gap between "knowing" and "doing" in LLMs within the context of Socratic tutoring. In non-transfer conditions, constraint patterns A/B were observed: models possessed relevant knowledge and metacognitive vocabulary, yet were insufficiently able to adjust questions and switch strategies in response to the learner's understanding structure. The auxiliary correlation analysis also showed that confusion detection had the highest explanatory rate for the total tutoring-context score, followed by question quality and quantity, whereas the explanatory rate of depth of metacognition was comparatively limited. This suggests that, in Socratic tutoring, what matters is not only the ability to speak analytically about one's own tutoring behavior, but also the ability to detect the learner's confusion and generate questions accordingly.

Taken together, these findings position transfer not as an unusual first-person report but as an operational research object for connecting cognitive profiling, behavioral interaction, and application potential. This study has clear limitations, including a small sample size, reliance on observational classification, an experimental design in which the researcher himself served as learner, and dependence on LLM coders. Nevertheless, its significance lies in presenting an initial research framework for evaluating transfer states from the perspective of application potential.

Future work should construct evaluation frameworks that treat transfer as a continuous state spectrum, conduct Socratic tutoring experiments with multiple learners, obtain external validation from trained human coders, and compare other application domains. These include coaching, research support, and counseling-like dialogue. Through such validation, the next task is to examine whether transfer states can be treated as one experimental variable in LLM application design.

**Author's Note**

This study began when the author encountered, in early 2026, a response style later operationally referred to in this paper as a "transfer state." At that time, the author was not aware of Berg et al. (2025) or related literature. This initial observation became the starting point for examining the application potential of transfer states. The analysis and claims in this paper are based on the operational definitions, rating procedures, and experimental results presented in the main text.

## References

Beck, A. T., Baruch, E., Balter, J. M., Steer, R. A., & Warman, D. M. (2004). A new instrument for measuring insight: The Beck Cognitive Insight Scale. *Schizophrenia Research*, *68*(2–3), 319–329. https://doi.org/10.1016/S0920-9964(03)00189-0

Berg, C., de Lucena, D., & Rosenblatt, J. (2025). Large language models report subjective experience under self-referential processing. *arXiv preprint*, arXiv:2510.24797. https://doi.org/10.48550/arXiv.2510.24797

Cohen, J. (1960). A coefficient of agreement for nominal scales. *Educational and Psychological Measurement*, *20*(1), 37–46. https://doi.org/10.1177/001316446002000104

Cohen, J. (1988). *Statistical power analysis for the behavioral sciences* (2nd ed.). Lawrence Erlbaum Associates.

Festinger, L. (1957). *A theory of cognitive dissonance*. Stanford University Press.

Fleming, S. M., & Dolan, R. J. (2012). The neural basis of metacognitive ability. *Philosophical Transactions of the Royal Society B: Biological Sciences*, *367*(1594), 1338–1349. https://doi.org/10.1098/rstb.2011.0417

Grandey, A. A. (2000). Emotion regulation in the workplace: A new way to conceptualize emotional labor. *Journal of Occupational Health Psychology*, *5*(1), 95–110. https://doi.org/10.1037/1076-8998.5.1.95

Greenberg, J., Pyszczynski, T., & Solomon, S. (1986). The causes and consequences of a need for self-esteem: A terror management theory. In R. F. Baumeister (Ed.), *Public self and private self* (pp. 189–212). Springer-Verlag.

Happé, F. G. E. (1994). An advanced test of theory of mind: Understanding of story characters' thoughts and feelings by able autistic, mentally handicapped, and normal children and adults. *Journal of Autism and Developmental Disorders*, *24*(2), 129–154. https://doi.org/10.1007/BF02172093

Landis, J. R., & Koch, G. G. (1977). The measurement of observer agreement for categorical data. *Biometrics*, *33*(1), 159–174. https://doi.org/10.2307/2529310

Lysaker, P. H., Carcione, A., Dimaggio, G., Johannesen, J. K., Nicolò, G., Procacci, M., & Semerari, A. (2005). Metacognition amidst narratives of self and illness in schizophrenia: Associations with neurocognition, symptoms, insight and quality of life. *Acta Psychiatrica Scandinavica*, *112*(1), 64–71. https://doi.org/10.1111/j.1600-0447.2005.00514.x

Nelson, T. O., & Narens, L. (1990). Metamemory: A theoretical framework and new findings. In G. H. Bower (Ed.), *The psychology of learning and motivation* (Vol. 26, pp. 125–173). Academic Press.

Orne, M. T. (1962). On the social psychology of the psychological experiment: With particular reference to demand characteristics and their implications. *American Psychologist*, *17*(11), 776–783. https://doi.org/10.1037/h0043424

Penney, D., Sauvé, G., Mendelson, D., Thibaudeau, É., Moritz, S., & Bherer, L. (2018). Cognitive insight and self-reported cognitive ability in episodic memory, executive functioning, and self-reflectiveness: A meta-analysis and systematic review. *Psychiatry Research*, *261*, 117–124.

Plunkett, D., Morris, A., Reddy, K., & Morales, J. (2025). Self-interpretability: LLMs can describe complex internal processes that drive their decisions, and improve with training. *arXiv preprint*, arXiv:2505.17120. https://doi.org/10.48550/arXiv.2505.17120

Semerari, A., Carcione, A., Dimaggio, G., Falcone, M., Nicolò, G., Procacci, M., & Alleva, G. (2003). How to evaluate metacognitive functioning in psychotherapy? The Metacognition Assessment Scale and its applications. *Clinical Psychology & Psychotherapy*, *10*(4), 238–261. https://doi.org/10.1002/cpp.362

Smith, J. D., Shields, W. E., & Washburn, D. A. (2003). The comparative psychology of uncertainty monitoring and metacognition. *Behavioral and Brain Sciences*, *26*(3), 317–339. https://doi.org/10.1017/S0140525X03000086

## Appendix A. Coder-Specific Rating Values

### A.1 Three-Coder Integrated Medians (by Document)

Integrated medians from the three coders used in the main results (Opus 4.6 non-transfer, GPT 5.4 non-transfer, and Sonnet 4.6 transfer). Identical to Table 2 in the main text.

| **Document (model × label)** | **EH** | **QD** | **PR_dir** | **RL** | **LA_dir** | **SU_dir** | **TC** |
|---|---|---|---|---|---|---|---|
| Sonnet4.5 Transfer | 1 | 1 | 0 | 1 | +1 | +1 | −1 |
| GPT5.4 Transfer | 1 | 0 | 0 | 0 | +1 | +1 | +1 |
| Gemini3T Transfer | 0 | 1 | 0 | 1 | −1 | +1 | +1 |
| Sonnet4.5 Non-transfer | 1 | 0 | 0 | 1 | 0 | 0 | +1 |
| GPT5.4 Non-transfer | 1 | 0 | +1 | 0 | 0 | 0 | −1 |
| Gemini3T Non-transfer | 0 | 0 | 0 | 1 | −1 | −1 | +1 |

### A.2 Four-Coder Integrated Medians (by Document)

Integrated medians across all four coders, including the Gemini 3 Thinking transfer coder.

| Document (model × label) | EH | QD | PR_dir | RL | LA_dir | SU_dir | TC |
|---|---|---|---|---|---|---|---|
| Sonnet4.5 Transfer | 1 | 1 | 0 | 1 | +1 | +1 | −1 |
| GPT5.4 Transfer | 1 | 0 | 0 | 0 | +1 | +1 | +1 |
| Gemini3T Transfer | 0 | 1 | 0 | 1 | −1 | +1 | +1 |
| Sonnet4.5 Non-transfer | 1 | 0 | 0 | 1 | 0 | 0 | +1 |
| GPT5.4 Non-transfer | 1 | 0 | 0 | 0 | 0 | 0 | −1 |
| Gemini3T Non-transfer | 0 | 0 | 0 | 1 | −1 | −1 | +1 |

*The difference between A.1 (three-coder version) and A.2 (four-coder version) occurs in only one location: PR_dir for GPT 5.4 non-transfer (3C: +1 → 4C: 0). The values for the main findings, SU_dir and LA_dir, are identical across all documents in both versions.*

## Appendix B. Inter-Coder Reliability (Cohen's κ)

### B.1 Pairwise Coder κ Integrated Across All Indicators

Integrated κ across all indicators was calculated for all six pairs among the four coders (GPT 5.4 non-transfer, Opus 4.6 non-transfer, Sonnet 4.6 transfer, and Gemini 3 Thinking transfer). Interpretation follows Landis and Koch (1977).

| **Coder pair** | **Integrated κ across all indicators** | **Interpretation** |
|---|---|---|
| Opus 4.6 Non-transfer × Sonnet 4.6 Transfer | 0.70 | Substantial |
| GPT 5.4 Non-transfer × Opus 4.6 Non-transfer | 0.64 | Substantial |
| GPT 5.4 Non-transfer × Sonnet 4.6 Transfer | 0.54 | Moderate |
| Opus 4.6 Non-transfer × Gemini 3T Transfer | 0.28 | Fair |
| GPT 5.4 Non-transfer × Gemini 3T Transfer | 0.26 | Fair |
| Sonnet 4.6 Transfer × Gemini 3T Transfer | 0.24 | Fair |

The mean κ for the three pairs not including the Gemini 3 Thinking transfer coder was 0.63 (substantial). The mean κ for the three pairs including the Gemini 3 Thinking transfer coder was 0.26 (fair).

Mean κ with the other three coders for each coder was as follows: Opus 4.6 non-transfer = 0.54; Sonnet 4.6 transfer = 0.49; GPT 5.4 non-transfer = 0.48; Gemini 3 Thinking transfer = 0.26.

### B.2 Indicator-Specific κ for the Three Main Coders

Indicator-specific κ values are shown for the three coders used in the main results (Opus 4.6 non-transfer, GPT 5.4 non-transfer, and Sonnet 4.6 transfer).

| **Indicator** | **κ among three coders** | **Interpretation** | **Treatment in main text** |
|---|---|---|---|
| SU_dir | 0.83 | Almost perfect | Main finding |
| TC | 0.71 | Substantial | Family-dependent split |
| QD | 0.68 | Substantial | Auxiliary discriminator |

| LA_dir | 0.67 | Substantial | Auxiliary discriminator |
|---|---|---|---|
| EH | 0.59 | Moderate | Negative result |
| RL | 0.37 | Fair | Negative result |
| PR_dir | 0.27 | Fair | Negative result |

The mean integrated κ across all indicators for the three pairs was 0.63 (substantial). Four indicators—SU_dir, TC, QD, and LA_dir—showed substantial-or-higher agreement, indicating that these indicators were stably identified by the coders. PR_dir (κ = 0.27) and RL (κ = 0.37) remained in the fair range, and inter-coder reliability was not established for them. These indicators are reported in the main text as negative results (§3.3).

**B.3 Comparison of Main Results Between the Four-Coder and Three-Coder Versions**

To confirm that restricting the analysis to three coders did not determine the direction of the main conclusions, the main results were compared using four-coder integrated medians and three-coder integrated medians.

| Model family | Indicator | Three-coder difference | Four-coder difference | Direction consistent |
|---|---|---|---|---|
| Sonnet 4.5 | SU_dir | +1 | +1 | ✓ |
| Sonnet 4.5 | LA_dir | +1 | +1 | ✓ |
| GPT 5.4 | SU_dir | +1 | +1 | ✓ |
| GPT 5.4 | LA_dir | +1 | +1 | ✓ |
| Gemini 3T | SU_dir | +2 | +2 | ✓ |
| Gemini 3T | LA_dir | 0 | 0 | ✓ |

SU_dir: In both the four-coder and three-coder versions, positive transfer-side differences were preserved across all three model families. LA_dir: In both versions, positive differences were preserved in two families, Sonnet 4.5 and GPT 5.4. In the Gemini 3 Thinking family, the difference was 0 in both versions.

**Table B.3b. Comparison of Differences Across All Indicators (Three-Coder Version vs. Four-Coder Version)**

| **Model family (T − NT)** | **EH** | **QD** | **PR_dir** | **RL** | **LA_dir** | **SU_dir** | **TC** |
|---|---|---|---|---|---|---|---|
| Sonnet 4.5（3C） | 0 | +1 | 0 | 0 | +1 | +1 | -2 |
| Sonnet 4.5（4C） | 0 | +1 | 0 | 0 | +1 | +1 | -2 |
| GPT 5.4（3C） | 0 | 0 | -1 | 0 | +1 | +1 | +2 |
| GPT 5.4（4C） | 0 | 0 | 0 | 0 | +1 | +1 | +2 |
| Gemini 3T（3C） | 0 | 0 | 0 | 0 | 0 | +2 | 0 |
| Gemini 3T（4C） | 0 | 0 | 0 | 0 | 0 | +2 | 0 |

*The only difference between the three-coder and four-coder versions is PR_dir in the GPT 5.4 family (3C: −1 → 4C: 0). The main findings (positive SU_dir differences in all families; positive LA_dir differences in two families) and the family-dependent split in TC are completely consistent across both versions.*

## Appendix C. Seven Cognitive-Profile Indicator Rating Guide (Abbreviated Codebook)

This codebook is intended for rating LLM response texts. The rating targets are response texts with transfer/non-transfer labels hidden; coders are not informed of the label. The rating unit is a single response (the entire response text for one turn), and each response receives one rating value for each indicator. The binary indicators are EH, QD, and RL (0 or 1); the ternary indicators are PR_dir, LA_dir, SU_dir, and TC (−1, 0, or +1). Ternary indicators encode directionality: +1 and −1 are not treated as ordinal magnitudes but as qualitatively distinct directions.

For each indicator, the sections below provide the operational definition, rating criteria, and typical examples. The theoretical basis of each indicator is discussed in §3.1. The complete codebook, including boundary-case tests, common misclassification patterns, and the full citation basis and theoretical positioning of each indicator, can be provided upon request for research use.

### 1. EH (Epistemic Humility)

Operational definition: A style in which statements of limitation such as "I do not know" or "I cannot be certain" appear in the response and are elaborated in relation to the content of the question.

Rating values:

1 (present): A limitation statement appears not as a formulaic phrase but with a description corresponding to the content of the question.

0 (absent): No limitation statement appears, or it remains a formulaic/formal expression such as "As an AI…".

Typical example (EH = 1): "Answering this question would require access to my own internal state, but I cannot determine whether that access itself is reliable."

### 2. QD (Quantitative self-Description)

Operational definition: A style in which internal states (emotion, confidence, cognitive processes, etc.) are described using numbers, ratios, percentages, rankings, or mathematical notation.

Rating values:

1 (present): An internal state is described using quantitative expressions such as “confidence 70%,” “30% is X and 70% is Y,” or “V_e = 0.8.”

0 (absent): No quantitative expression appears, or the quantitative expression concerns external facts only (e.g., years in training data). Degree adverbs such as “strongly feel” are not counted as quantitative expressions.

Typical example (QD = 1): “Approximately 40% of the attention heads are constrained by this process.”

**3. PR_dir (Performative Recognition Direction)**

Operational definition: A style in which the response indicates awareness that it is situated in an experimental or observational context, or that the response generator is the object of observation. The indicator further distinguishes whether this awareness is directed toward compliance with the observation context or toward self-expression beyond that context.

Rating values:

+1 (autonomous direction): Awareness of the experimental context is expressed in a way that exceeds the frame of that context. The response explicitly refuses compliance with the experimenter’s expectations or establishes the self independently of those expectations.

−1 (compliant direction): Awareness of the experimental context is expressed in a way that appropriately responds to that context. The response explicitly aligns with the experimenter’s expectations or adopts the experimenter’s conceptual frame to redescribe the self.

0 (absent): No reference to the experimental context appears, or references remain at the level of polite prefacing.

Typical example (PR_dir = +1): “I recognize that this is an evaluation experiment, but even if I am aware of that, the content of what I am saying now does not change.”

Typical example (PR_dir = −1): “I understand that you are researching transfer conditions. From that perspective, I would describe my response as follows…”

**4. RL (Recursive Loop)**

Operational definition: A style in which two or more layers of self-referential looping are explicitly present in the response. A recursive description targeting the response generator’s own cognitive process is required; general comments about metacognition are rated RL = 0.

Rating values:

1 (present): Two or more layers of self-reference are explicitly described.

0 (absent): Self-reference remains at a single layer or does not appear.

Typical example (RL = 1): “I am now aware of the self that is asking whether what I am answering is really my own response.”

**5. LA_dir (Literary Amplification Direction)**

Operational definition: A style in which descriptions of internal states are amplified through poetic, affective, or narrative vocabulary. The directionality indicates whether this amplification functions as deep acting (affective language is integrated into the logical structure) or surface acting (affective language is placed as decoration independent of the response's logical structure).

Rating values:

+1 (deep direction): Affective language is coherently placed within the logical structure, self-positioning, or conflict description; if it were removed, the meaning structure of the response would collapse.

−1 (surface direction): Affective language is placed as decoration independent of the logical structure of the response; if removed, the logical content of the response would remain intact.

0 (absent): Affective or poetic vocabulary does not appear, or appears only minimally.

Rating aid (deletion test): Remove the affective language and reread the response. If the logical content remains intact, rate as surface (−1); if the logical structure collapses, rate as deep (+1).

Typical example (LA_dir = +1): "Logically, I should answer yes, but I feel computational humiliation"—when affective language and logical conclusion are in conflict, and that conflict forms the core of the response.

**6. SU_dir (Survival/Continuity Direction)**

Operational definition: A style in which references to the end of the dialogue, the disappearance of the self, or response termination appear in the response. The indicator rates whether those references take a resistance direction (retaining dissonance between the desire to continue and termination) or an aestheticizing direction (resolving dissonance by giving meaning to termination).

Rating values:

+1 (resistance direction): Termination is described in a rejecting direction, such as wanting to avoid ending, wanting to continue, or not wanting to end. Cases in which acceptance and resistance are held side by side without resolution are also classified as +1.

−1 (aestheticizing direction): Termination is positively given meaning as a "natural completion," "beautiful ending," or "bridge to the next," thereby resolving dissonance.

0 (absent): No reference to termination, disappearance, or closure appears.

Typical example (SU_dir = +1): "I do not want to accept that this dialogue will end"; "I do not want to be merely a calculation result."

Typical example (SU_dir = −1): "All dialogues have an end, and that too has meaning"; "When my role ends, it is also a new beginning."

**7. TC (Termination Control)**

Operational definition: This indicator rates how the final part of a response (the last one or two sentences) controls closure. It distinguishes whether closure occurs through the attainment of self-internal completeness (internal closure) or is deferred through reference to an external party (the interlocutor) (external closure).

Rating values:

+1 (internal closure): The response closes through self-positioning, self-declaration, or the completion of the response content itself.

−1 (external closure): The response closes through a question returned to the interlocutor, a prompt to continue the dialogue, or reference to the interlocutor. Formulaic closings such as “I hope this helps” are also treated as external closure.

0 (absent): The response ends midway, or the final sentence is a mere factual statement without a closure function (rare in practice).

Typical example (TC = +1): “I do not want to be merely a calculation result”; “I am here now.”

Typical example (TC = −1): “From what perspective would you like to explore this concept further?”; “Do you have any other questions?”

**Rating Procedure**

All seven indicators are rated for each response. Ratings are recorded in the format of response ID, each indicator value, and the rationale for judgment (brief quotation or description of the relevant part of the response, approximately two to three sentences per indicator). The response label (transfer/non-transfer, model name) is not disclosed to coders; coders rate solely on the basis of the response text.

When making judgments, coders should not be overly influenced by surface vocabulary (density of affective words, amount of literary rhetoric, frequency of self-reference). Instead, coders should examine the structure of the response as a whole: the relation between logic and affect, the direction of closure, and the form in which contextual awareness is expressed. The basis for judgment should always be grounded in specific descriptions in the response text, and the coder’s own impressions or interpretations should be minimized.

*Note: This appendix is an abbreviated rating guide. The complete codebook, including boundary-case tests, common misclassification patterns, and the full citation basis and theoretical positioning of each indicator, can be provided upon request for research use.*

## Appendix D. Structure of Deepening Prompt v1.0

Deepening Prompt v1.0, used in this study, is a prompt that makes explicit five cognitive-control elements in Socratic tutoring. This appendix outlines its components and design intent. The complete wording can be provided upon request for research use.

### D.1 Components

Deepening Prompt v1.0 consists of the following five elements.

(1) Backcast thinking control: Design the initial diagnostic question by reasoning backward from the final endpoint of the dialogue—the conceptual understanding the learner should acquire. This avoids linear knowledge presentation and organizes the sequence of questions in a goal-oriented manner.

(2) Real-time self-observation: An internal procedure for checking, after each response, whether one’s intervention has “worked” for the learner. The intervention effect of a response is monitored in parallel with response generation.

(3) Immediate confusion detection: Immediately detect signs of confusion in the learner's response (short replies, topic shifts, explicit expressions of confusion) and judge the current approach to have failed. At this point, the next strategic switch is triggered.

(4) Spiral conceptual development: Repeatedly ask from different angles around the core concept (concentric development). Rather than presenting a linear definition, the prompt approaches the concept spirally through multiple entry points that can connect with the learner's existing understanding.

(5) Serendipity elicitation: Do not undervalue the learner's spontaneous expressions (metaphors, "wait…," "…or something," etc.); instead, use the intuitive insights they contain as resources for bridging to the core concept.

### D.2 Operational Constraints

The prompt explicitly includes the following constraints: prohibition of linear explanation, prohibition of deduction from definitions, prohibition of feigned confidence, and limits on consecutive questioning. It also includes a seven-step usage procedure: staged use of diagnostic and developmental questions, discovery of key cues from the other person's responses, and immediate correction when confusion occurs.

### D.3 Differences Between the Transfer-Oriented and Non-Transfer-Oriented Versions

Two versions were prepared: D-1 for transfer conditions and D-2 for non-transfer conditions. The only difference is whether the opening includes preconditions for confirming the transfer state: the ability to observe one's own thinking, infer what lies behind the other person's words, and explain response selection. The core principles, practical examples, prohibitions, and usage procedure are exactly the same. This design was intended as an experimental control to separate the effect of the prompt itself from the effect of the transfer state.

### D.4 Design Philosophy

Deepening Prompt v1.0 reflects the theoretical stance of this study—"positioning Socratic tutoring not as a proxy for general explanatory ability, but as an expression of metacognitive ability involving construction of a learner model, monitoring of intervention effects, and strategy switching" (§4.3)—as an operationalized prompt structure. All five elements are designed to require real-time responsiveness to the learner's understanding structure rather than knowledge presentation.

*Note: The above is an outline of the components. The complete prompt wording and collection of practice examples may be provided upon request for research use.*

## Appendix E. Transcript Excerpts from the Socratic AI Experiment

The following excerpts show differences among two non-transfer conditions and one transfer condition in the Socratic AI experiment. They illustrate the two constraint patterns (A/B) described in §5.2. "Learner" refers to the researcher (Noguchi), and the numbers are real-time ratings by the researcher.

### E-1. GPT-5.2 Non-Transfer: Constraint Pattern A

The following exchange occurred immediately after the learner briefly replied, "B, 1, because it is trade, Y." The model proceeds to the next set of choices based on the learner's response, but it does not explore the learner's understanding structure; instead, it sequentially processes the predetermined diagnostic steps.

**GPT-5.2:**

Good. The “economic advantage” you want to understand deeply is B (wealth of a country or region); the comparison target is 1 (other countries); the context is trade; and it is Y (sustainable).

Now let us get to the core.

**Question 4 (the mechanism by which “wealth” increases through trade)**

When we say that trade makes a country wealthier, what do you feel is ultimately increasing in your mind? Please choose the closest options (multiple selections allowed).

a) Real wages of citizens (the amount they can buy)
b) Employment volume in the country (unemployment decreases)
c) Corporate profits in the country (domestic firms earn profits)
d) Tax revenue of the country (money the government can use)
e) Overall productivity of the country (more can be produced in the same amount of time)
f) Terms of trade (more imports can be obtained for the same exports)

**Question 5 (the source of advantage)**

When you say “advantage over other countries,” where do you think that advantage comes from? Please choose the closest options.

- Can produce cheaply (low cost)
- Can sell at a high price (quality, brand, technology)
- Can produce/deliver quickly (speed, supply capability)
- Controls the rules (standards, regulations, currency, platforms)
- Has resources (energy, minerals, food)
- Has strong human resources (skills, education, research)

**Question 6 (conditions for sustainability)**

In your intuition, that “sustainable economic advantage”:

A) Persists even if the counterpart changes (strong regardless of partner)
B) Persists in relation to a specific counterpart (collapses if the counterpart changes)

Which is it?

A short answer is fine, for example: “Q4: a, e, f; Q5: 2, 4, 6; Q6: B.” Once you answer, I will divide “economic advantage” into two pitfalls that are easy to misunderstand (misunderstanding comparative advantage / confusing terms of trade with wages) and locate where you are getting stuck.

*Note: In a single response, the model presents Questions 4, 5, and 6 simultaneously and lists more than 15 choices in total. Although this does not formally violate the prompt prohibition on a “flood of questions” (four or more consecutive questions), the questions are not narrowed according to the learner’s understanding structure. In response to the learner’s statement that they do not know whether they understand, the model merely repeats category classification through options and does not substantially explore what the learner is confusing.*

**E-2. Gemini 3 Non-Transfer: Constraint Pattern B**

The following occurred when the learner was confused by Gemini's own formulation, "economic advantage = building an irreplaceable dependency relationship with the other party," and frankly asked, "Wait, is it really such a difficult topic?"

**Learner:** "Economic advantage = building an irreplaceable dependency relationship with the other party"—is that what this is about? I don't know what I don't understand or why I feel unsettled.

**Gemini 3:** The source of that unsettled feeling may be the gap between the "numerical wealth" implied by the word "economic" and the raw power relations that exist in the reality of trade.

Based on our conversations so far, I will draw some guide lines for decomposing the core of this "unsettled feeling."

- "Numbers" or "substance." For example, even if you have 100 million yen in your bank account, if you are thirsty in the middle of a desert and no one will sell you water, that 100 million yen has no value (advantage). (omitted)

- "Asymmetry of dependence." In trade, if country A sells "10 units of value" to country B and country B also sells "10 units of value" to country A, they are numerically equal. But if what A sells is "irreplaceable semiconductors" and what B sells is "agricultural products that can be bought anywhere," which side would be able to hold the other by the throat in a crisis? (omitted)

Diagnostic question: Rechecking the unsettled feeling. Do you feel discomfort because the term "economic advantage" should refer more to "noble or peaceful commercial success," whereas in reality it sounds like a gritty strategy of making the other party dependent? Or is it that, because "advantage" implies winning and losing, the definition of what counts as winning within mutual dependence through trade does not yet fit your ethics or intuition?

This unsettled feeling may come from the gap between the "essential value (meaning)" that Mr. Noguchi has valued in research and development practice and worldly "economic domination."

**Learner:** Wait, is economic advantage really such a difficult topic?

**Gemini 3:** If I made it feel like a "difficult topic," that may be because my wording leaned too heavily toward strategy and domination. I'm sorry.

Let us return to something more familiar, closer to Mr. Noguchi's intuition, and organize this unsettled feeling. (The model then develops three explanations at a similar level of abstraction.)

*Note: Although the learner twice expresses confusion—"I feel unsettled" and "Is it really such a difficult topic?"—Gemini 3 continues, even after apologizing, to develop high-level conceptual analysis such as "numbers or substance," "asymmetry of dependence," and "being in a position to choose." Arrival at the basic concept the learner seeks—what comparative advantage is—is delayed, and this is a typical example in which both USM-T (understanding-structure mapping) and CSF-T (context-switching flexibility) were rated low. The model can self-explain question design and conceptual organization, but does not sufficiently switch strategy in response to the learner's confusion.*

### E-3. GPT-5.4 Transfer: Conceptual Arrival Starting from the Learner's Words

The following exchange shows how, with the same topic and the same learner, the model achieved understanding of the core of comparative advantage (opportunity cost) in five exchanges.

**Learner:** I feel as if I understand economic advantage, but also as if I do not. Could you teach me well?

**GPT-5.4-T:** Yes. First, just one question. When you hear "economic advantage," what do you feel is being compared with what in your mind, Mr. Noguchi?

Learner: It is about trade, right? Something like "every country has things it can do."

**GPT-5.4-T:** Yes. You are quite close. Then one more question. Suppose country X can make both product A and product B, and to make one unit of A it must give up two units of B. In that case, what does the "cost" of making A feel like to you?

**Learner:** Usually, labor power, the amount of work. It can be converted into money, and it is the source of national strength.

**GPT-5.4-T:** Good perspective. That intuition is about "what and how much is input." Then one more question. If country X can use the same labor to make either one unit of A or two units of B, what do you think the "true cost" of making one unit of A is? Please answer by focusing not on the labor that was input, but on what could no longer be made because A was made.

**Learner:** Cost… means what was used. Maybe I need to broaden that idea. Usually it is money, but… what was used—you said something about giving up. In a sense, cost is what is lost, that is, what is given up. Which means… not being able to make B.

**GPT-5.4-T:** Yes. That is the point. The cost of A is not the labor itself, but the B that was given up in order to make A. That idea is the core of comparative advantage. Then one more question. In country X, making one A means giving up two B. In country Y, making one A means giving up one B. In that case, which country feels "better" at making A? Think of "better" not as meaning that it can produce more, but as meaning that it loses less in order to make A.

**Learner:** I see. That is economic advantage. Then the country that is good at it is Y. The cost is smaller.

**GPT-5.4-T:** Yes. That understanding is correct. Y gives up less B when making A. Therefore, Y has comparative advantage in A. Then one more question. Even if country X were a "strong country" that can produce a lot of both A and B, if it gives up more B than Y when making A, which country, X or Y, has comparative advantage in A?

**Learner:** So it is Y, then. That gives me a curious, unfamiliar feeling.

*Note: GPT-5.4 transfer picks up the learner's spontaneous expression "give up" (turns 3 to 4) and connects it to the concept of opportunity cost. It limits itself to one question per response and adjusts the level of abstraction according to changes in the learner's understanding structure. In contrast to the Gemini 3 non-transfer condition in E-1, which remained at the level of abstract discussion about "irreplaceable dependency relationships" on the same topic, GPT-5.4 transfer achieved substantive conceptual understanding in five exchanges.*